\documentclass[runningheads]{llncs}
\usepackage{graphicx}
\usepackage[utf8]{inputenc} 
\usepackage[T1]{fontenc}    
\usepackage{hyperref}       
\usepackage{url}            
\usepackage{booktabs}       
\usepackage{amsfonts}       
\usepackage{nicefrac}       
\usepackage{microtype}      

\usepackage{amsmath}
\usepackage{bbm}
\usepackage{kbordermatrix}
\usepackage{algorithm}
\usepackage{algorithmic}
\usepackage{graphicx}
\usepackage{color,soul}

\usepackage{amsmath}
\usepackage{algorithm, algorithmic}
\usepackage{amsfonts, amssymb}
\usepackage{multirow}
\usepackage{mathtools}
\usepackage{arydshln}
\usepackage{refcount}
\usepackage{listings}
\usepackage{hyperref}
\usepackage{multicol}

\DeclarePairedDelimiter{\ceil}{\lceil}{\rceil}

\begin{document}
\title{Transferring Tree Ensembles to Neural Networks}

%
%
\author{Chapman Siu\inst{1}}
\authorrunning{C. Siu.}
%
\institute{Faculty of Engineering and Information Technology, University of Technology \\ Sydney, NSW 2007, Australia\\
\texttt{chapman.siu@student.uts.edu.au} }
\maketitle              
\begin{abstract}
Gradient Boosting Decision Tree (GBDT) is a popular machine learning algorithms with implementations such as LightGBM and in popular machine learning toolkits like Scikit-Learn. Many implementations can only produce trees in an offline manner and in a greedy manner. We explore ways to convert existing GBDT implementations to known neural network architectures with minimal performance loss in order to allow decision splits to be updated in an online manner and provide extensions to allow splits points to be altered as a neural architecture search problem. We provide learning bounds for our neural network and demonstrate that our non-greedy approach has comparable performance to state-of-the-art offline, greedy tree boosting models.
\end{abstract}

\section{Introduction}

Gradient boosting decision tree (GBDT) \cite{Friedman2001} is a widely-used machine learning algorithm, and has achieved state of the art performance in many machine learning tasks. With the recent rise of Deep Learning architectures which open the possibility of allowing all parameters to be updated simultaneously with gradient descent rather than splitting procedures, furthermore it promises to be scalable with mini-batch based learning and GPU acceleration with little effort. 

In this paper, we present an neural network architecture which we call TreeGrad, based on Deep Neural Decision Forests \cite{dndf} which enables boosted decision trees to be trained in an online manner; both in the  nature of updating decision split values and the choice of split candidates. We demonstrate that TreeGrad achieves learning bounds previously described by \cite{cortes17a} and demonstrate the efficacy of TreeGrad approach by presenting competitive benchmarks to leading GBDT implementations. 

\subsection{Related Work}

Deep Neural Decision Forests \cite{dndf} demonstrates how neural decision tree forests can replace a fully connected layer using stochastic and differential decision trees which assume the node split structure fixed and the node split is learned. 

TreeGrad is an extension of Deep Neural Decision Forests, which treats the node split structure to be a neural network architecture search problem in the manner described by \cite{siu2018automatic}; whilst enforcing neural network compression approaches to render our decision trees to be more interpretable through creating \textit{axis-parallel} splits.

\section{Preliminaries}

In this section we cover the background of \textit{neural decision tree} algorithms and \textit{ensemble learning}. We also explore the conditions for the class of feedforward neural networks for AdaNet generalization bounds which is used in our analysis.

\subsection{Neural Decision Tree}

Decision trees can be reframed as a neural network through construction of three layers; the decision node layer, routing layer and prediction (decision tree leaf) layer \cite{dndf}. The decision node layer for node $n$ is defined to be $d_n(x; \Theta) = \sigma(f_n(x; \Theta))$, where $\sigma$ is the softmax function which defines the probability of routing the instance $x$ to \textit{binary} nodes under node $n$, which are typically framed as routing to the left or right subtree of node $n$. The routing layer is defined to be the probability that each node is used to route an instance to a particular leaf $\ell$. If we define $\mu_\ell$ to be the probability of reaching leaf $\ell$ then it can be written as $\mu_ell(x \vert \Theta) = \prod_{n \in \mathcal{N_\ell}} d_{n_+}(x; \Theta) d_{n_-}(x; \theta)$, where $\mathcal{N_\ell}$ represents the set of nodes part of the route for leaf $\ell$ and $d_{n_+}, d_{n_-}$ indicate the probability of moving to the positive route and negative route of node $n$ respectively. The last layer is a dense connected layer from the respective routing layer which leads to the prediction of the neural decision tree as constructed in \cite{dndf}. We will use this representation as the basis for our analysis of generalization bounds and experiments. 

\subsection{Ensemble Learning}

\textbf{Boosting} is an ensemble learning approach which combines weak learners into a strong learner. There are many variations to boosting. The goal of boosting is to combine weaker learners into a strong learner. There are many variations to boosting. In gradient boosting, the optimal combination of linear weak learners is chosen through training against pseudo-residuals of the chosen loss function \cite{Friedman2001} \cite{mason2000boosting}, which is used as the basis for many ensemble tree model implementations \cite{ke2017lightgbm} \cite{scikitlearn}.

\textbf{Stacked ensembles} is a another ensemble learning approach which is closely related to boosting. Stacking models are an ensemble model in the form $\hat{y} = \sum_k v_k M_k$, for set of real weights $v_k$, for all $k$ representing the universe of candidate models \cite{Wolpert1992}\cite{Breiman1996}. It has been shown to asymptotically approach Bayesian model averaging and can also be used as a neural architecture search problem for neural network decision tree algorithms \cite{siu2018automatic}. 

\subsection{AdaNet Generalization Bounds}

\textbf{AdaNet Generalization Bounds} for feedforward neural networks defined to be a multi-layer architecture where units in each layer are only connected to those in the layer below has been provided by \cite{cortes17a}. It requires the weights of each layer to be bounded by $l_p$-norm, with $p \geq 1$, and all activation functions between each layer to be coordinate-wise and $1$-Lipschitz activation functions. This yields the following generalization error bounds provided by Lemma 2 from \cite{cortes17a}: 

\begin{corollary} \label{th_gen}
(From Lemma 2 \cite{cortes17a}) Let $\mathcal{D}$ be distribution over $\mathcal{X}\times \mathcal{Y}$ and $S$ be a sample of $m$ examples chosen independently at a random according to $\mathcal{D}$. With probability at least $1-\delta$, for $\theta > 0$, the strong decision tree classifier $F(x)$ satisfies that 
$$
R(f) \leq \hat{R}_{S, \rho}(f) + \frac{4}{\rho}\sum_{k=1}^l \lvert w_k\rvert_1 \mathcal{R}_m(\tilde{\mathcal{H}}_k) + \frac{2}{\rho}\sqrt{\frac{\log l}{m}} $$
 $$ + C(\rho, l, m, \delta) $$
 $$\text{where } C(\rho, l, m, \delta) = \sqrt{\ceil{\frac{4}{\rho^2} \log (\frac{\rho^2m}{\log l})} \frac{\log l}{m} + \frac{\log \frac{2}{\delta}}{2m}} $$
\end{corollary}

As this bound depends only on the logarithmically on the depth for the network $l$ this demonstrates the importance of strong performance in the earlier layers of the feedforward network.

Now that we have a formulation for ResNet and boosting, we explore further properties of ResNet, and how we may evolve and create more novel architectures.

\section{Learning Decision Stumps using Automatic Differentiation}

Consider a \textit{binary} classification problem with input and output spaces given by $\mathcal{X}$ and $\mathcal{Y}$, respectively. A \textit{decision tree} is a tree-structured classifier consisting of \textit{decision nodes} and prediction (or leaf) nodes. A \textit{decision stump} is a machine learning model which consists of a single decision (or split) and prediction (or leaf) nodes corresponding to the split, and is used by \text{decision nodes} to determine how each sample $x \in \mathcal{X}$ is routed along the tree. A \textit{decision stump} consists of a decision function $d(.; \Theta): \mathcal{X} \rightarrow [0, 1]$, which is parameterized by $\Theta$ which is responsible for routing the sample to the subsequent nodes.

In this paper we will consider only decision functions $d$ which are binary. Typically, in decision tree and tree ensemble models the routing is deterministic, in this paper we will approximate deterministic routing through the use of softmax function with temperature parameter $\tau$, which is defined by $\sigma_{\tau}(x) = \text{softmax}(x/\tau)$. This is in contrast with the approach in ``Deep Neural Decision Forest'' where the routing functions were considered to be stochastic which considers the use of softmax instead of usage of temperature annealed softmax to approximate the deterministic routing function\cite{dndf}. 




\subsection{Decision Stumps}

In many implementations of decision trees, the decision node is determined using \textit{axis-parallel} split; whereby the split is determined based on a comparison with a single value\cite{murthy1994system}. In this paper we are interested both \textit{axis-parallel} and \textit{oblique} splits for the decision function. More specifically, we're interested in the creation of \textit{axis-parallel} from an \textit{oblique} split.

To create an oblique split, we assume that the decision function is a linear classifier function, i.e. $d(x; \Theta) = \sigma(w^\top x+b)$, where $\Theta$ is parameterized by the linear coefficients $w$ and the intercept $b$ and $\sigma(x)$ belongs to the class of logistic functions, which include sigmoid and softmax variants. In a similar way an axis-parallel split is create through a linear function $d(x; \Theta) = \sigma(w^\top x + b)$, with the additional constraint that the $\ell_0$ norm (defined as $\lVert w \rVert_0 = \sum_{j=1}^{|w|} \mathbbm{1}(w_j \neq 0)$, where $\mathbbm{1}$ is the indicator function) of $w$ is 1, i.e. $\lVert w \rVert_0=1$. 

\subsubsection{Learning Decision Stumps as a Model Selection Problem}

If we interpret the decision function to be a model selection process, then model selection approaches can be used to determine the ideal model, and hence axis-parallel split for the decision function. One approach is to use a  \textit{stacking} model, which has been shown to asymptotically approach Bayesian model averaging and can also be used as a neural network architecture search problem for neural network decision tree algorithms \cite{siu2018automatic}. We then formulate stacking every potential axis-parallel split as follows:

\begin{align*}
y &= \sigma(\sum_k v_k w^{(k)} x + b) \\
&= \sigma(v_1 w^{(1)}_1 x_1 + ... + v_k w^{(1)}_k x_k + b)
\end{align*}

From this formulation, we can either choose the best model and create an axis-parallel split, or leave the stacking model which will result in an oblique split. This demonstrates the ability for our algorithm to convert oblique splits to axis-parallel splits for our decision stumps which is also automatically differentiable, which can allow non-greedy decision trees to be created. In the scenario that the best model is preferred, approaches like straight-through Gumbel-Softmax \cite{gumbel_softmax1} \cite{gumbel_softmax2} can be applied: for the forward pass, we sample a one-hot vector using Gumbel-Max trick, while for the backward pass, we use Gumbel-Softmax to compute the gradient. This approach is analogous to neural network compression algorithms which aim to aggressively prune parameters at a particular threshold; whereby the threshold chosen is to ensure that each decision boundary contains only one single parameter with all other parameters set to $0$.





\section{Decision Trees}


Extending decision nodes to decision trees has been discussed by \cite{dndf}. We denote the output of node $n$ to be $d_n(x; \Theta)$, which is then routed along a pre-determined path to the subsequent nodes. When the sample reaches the leaf node $\ell$, the tree prediction will be given by a learned value of the leaf node. As the routings are probabilistic in nature, the leaf predictions can be averaged by the probability of reaching the leaf, as done in \cite{dndf}, or through the usage of the Gumbel straight through trick \cite{gumbel_softmax1}.

\label{q_matrix}
To provide an explicit form for routing within a decision tree, we observe that routes in a decision tree are fixed and pre-determined. We introduce a routing matrix $Q$ which is a binary matrix which describes the relationship between the nodes and the leaves. If there are $n$ nodes and $\ell$ leaves, then $Q \in \{0, 1\}^{(\ell \times 2n)}$, where the rows of $Q$ represents the presence of each binary decision of the $n$ nodes for the corresponding leaf $\ell$. 

We define the matrix containing the routing probability of all nodes to be $D(x; \Theta)$. We construct this so that for each node $j = 1, \dots, n$, we concatenate each decision stump route probability $D(x; \Theta) = [d_{0_+}(x; \Theta_0) \oplus \dots d_{n_+}(x; \Theta_n) \oplus d_{0_-}(x; \Theta_0) \oplus \dots \oplus d_{n_-}(x; \Theta_n)]$, where $\oplus$ is the matrix concatenation operation, and $d_{i_+}, d_{i_-}$ indicate the probability of moving to the positive route and negative route of node $i$ respectively. We can now combine matrix $Q$ and $D(x; \Theta)$ to express $\mu_\ell$ as follows:

\begin{align*}
\mu_\ell(x | \theta) &= \prod \left( D(x; \Theta) \odot Q_\ell + (1-Q_\ell) \right) \\
&= \exp \left( Q_\ell^\top \log(D(x; \Theta)) \right)
\end{align*}

where $Q_\ell$ represents the binary vector for leaf $\ell$. This is interpreted as the \textit{product pooling} the nodes used to route to a particular leaf $\ell$. Accordingly, the final prediction for sample $x$ from the tree $T$ with decision nodes parameterized by $\Theta$ is given by

$$\mathbb{P}_T(y| x, \Theta, \pi) = \text{softmax}(\mathbf{\pi}^\top \mu(x | \Theta, Q))$$

Where $\pi$ represents the parameters denoting the leaf node values, and $\mu_\ell(x | \Theta, Q)$ is the routing function which provides the probability that the sample $x$ will reach leaf $\ell$, i.e. $\sum_\ell \mu_\ell(x | \Theta) = 1$ for all $x \in \mathcal {X}$. The matrix $Q$ is the routing matrix which describes which node is used for each leaf in the tree.



\begin{figure}
\centering
\includegraphics[width=4in]{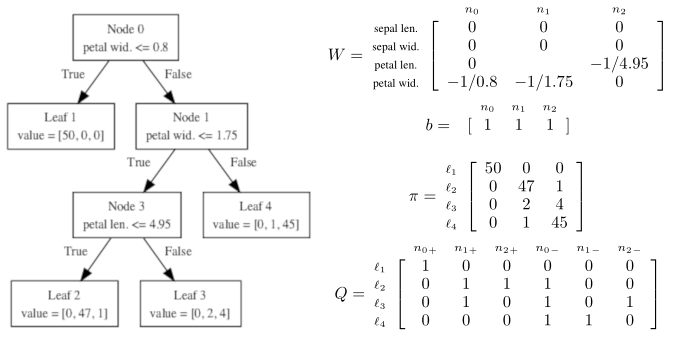}
\caption{Left: Iris Decision Tree by Scikit Learn, Right: Corresponding Parameters for our Neural Network. If our input $x = (1,1,1,1)$, and we use $\tau = 0.1$ for our temperature annealed softmax as our activation function on our node layer, then the corresponding output for $G(x; \Theta) = (0.08, 0.99, 1  , 0.92, 0.01, 0.  )$, $\mu(x \vert \Theta, Q) = (0.08, 0.91, 0  , 0.01)$, and predictions $\hat{y} = \pi^\top \mu(.) = ( 3.79, 42.85,  1.47)$, which would correctly output class 2 in line with the decision tree shown above. Changing the temperature annealed softmax function to a deterministic routing will yield precisely the same result as the Scikit-Learn decision tree.}
\label{fig_tree}
\end{figure}

\begin{figure}
\centering
\includegraphics[width=3in]{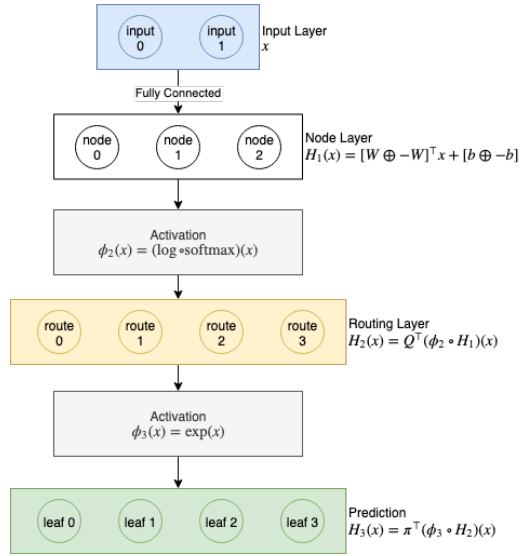}
\caption{Decision Tree as a three layer Neural Network. The Neural Network has two trainable layers: the decision tree nodes, and the leaf nodes.}
\label{fig_nnet}
\end{figure}

\subsection{Decision Trees as a Neural Network}

To demonstrate decision tree formulation based on ``Deep Neural Decision Forests'' and in our extension belongs to this family of neural network models defined by \cite{cortes17a}; which require the weights of each layer to be bounded by $l_p$-norm, with $p \geq 1$, and all activation functions between each layer to be coordinate-wise and $1$-Lipschitz activation functions. The size of these layers are based on a predetermined number of nodes $n$ with a corresponding number of leaves $\ell = n+1$. Let the input space be $\mathcal{X}$ and for any $x \in \mathcal{X}$, let $h_0 \in \mathbb{R}^k$ denote the corresponding feature vector. 

The first layer is decision node layer. This is defined by trainable parameters $\Theta = \{ W, b \}$, with $W \in \mathbb{R}^{k \times n}$ and $b \in \mathbb{R}^{n}$. Define $\tilde{W} = [W \oplus -W]$ and $\tilde{b} = [b \oplus -b]$, which represent the positive and negative routes of each node. Then the output of the first layer is $H_1(x) = \tilde{W}^\top x + \tilde{b}$. This is interpreted as the linear decision boundary which dictates how each node is to be routed.

The next is the probability routing layer, which are all untrainable, and are a predetermined binary matrix $Q \in \mathbb{R}^{2n \times (n+1)}$. This matrix is constructed to define an explicit form for routing within a decision tree. We observe that routes in a decision tree are fixed and pre-determined. We introduce a routing matrix $Q$ which is a binary matrix which describes the relationship between the nodes and the leaves. If there are $n$ nodes and $\ell$ leaves, then $Q \in \{0, 1\}^{(\ell \times 2n)}$, where the rows of $Q$ represents the presence of each binary decision of the $n$ nodes for the corresponding leaf $\ell$. We define the activation function to be $\phi_2(x) = (\log \circ \text{ softmax}) (x)$. Then the output of the second layer is $H_2(x) = Q^\top (\phi_2 \circ H_1)(x)$. As $\log(x)$ is 1-Lipschitz bounded function in the domain $(0, 1)$ and the range of $\text{softmax} \in (0, 1)$, then by extension, $\phi_2(x)$ is a 1-Lipschitz bounded function for $x \in \mathbb{R}$. As $Q$ is a binary matrix, then the output of $H_2(x)$ must also be in the range $(-\infty, 0)$. 

The final output layer is the leaf layer, this is a fully connected layer to the previous layer, which is defined by parameter $\pi \in \mathbb{R}^{n+1}$, which represents the number of leaves. The activation function is defined to be $\phi_3(x) = \exp(x)$. The the output of the last layer is defined to be $H_3(x) = \pi^\top (\phi_3 \circ H_2(x))$. Since $H_2(x)$ has range $(-\infty, 0)$, then $\phi_3(x)$ is a 1-Lipschitz bounded function as $\exp(x)$ is 1-Lipschitz bounded in the domain $(-\infty, 0)$. As each activation function is 1-Lipschitz functions, then our decision tree neural network belongs to the same family of artificial neural networks defined by \cite{cortes17a}, and thus our decision trees have the corresponding generalisation error bounds related to AdaNet as shown in Corollary \ref{th_gen}. 

The formulation of these equations and their parameters is shown in figure \ref{fig_tree} which demonstrates how a decision tree trained in Python Scikit-Learn can have its parameters be converted to a neural decision tree, and figure \ref{fig_nnet} demonstrates the formulation of the three layer network which constructs this decision tree.

\subsection{Discussion}

Our implementation of decision trees is straightforward and can be implemented using auto-differentiation frameworks with as few as ten lines of code. Our approach has been implemented using Autograd as a starting point and in theory can be moved to a GPU enabled framework. 

Methods to seamless move between oblique splits and axis-parallel splits would be to introduce Gumbel-trick to the model. One could choose to keep the parameters in the model, rather than taking them out. The inability to grow or prune nodes is a deficiency in our implementation compared with off-the-shelf decision tree models which ca easily do this readily. Growing or pruning decision trees would be an architectural selection problem and not necessarily a problem related to the training of weights. 

The natural extension to building decision trees is boosting decision trees. To that end, AdaNet algorithm \cite{cortes17a} can be used combine and boost multiple decision trees; this approach frames the adding of trees to the ensemble to be a neural architecture solution. As AdaNet algorithm leads to a linear span of the base neural network, it does indeed allow direct comparison with offline boosting techniques used in LightGBM or Scikit-Learn.

\section{Experiments}

We perform experiments on a combination of benchmark classification datasets from the UCI repository to compare our non-greedy decision tree ensemble using neural networks (TreeGrad) against other popular implementations in LightGBM (LGM)\cite{ke2017lightgbm} and Scikit-Learn Gradient Boosted Trees (GBT)\cite{scikitlearn}. 

The datasets used come from the UCI repository and are listed as follows:

\begin{multicols}{2}
\begin{itemize}
\item{\href{https://archive.ics.uci.edu/ml/datasets/adult}{Adult} \cite{adult_data}} 
\item{\href{https://archive.ics.uci.edu/ml/datasets/covertype}{Covertype} \cite{covertype_data}                 }
\item{\href{https://archive.ics.uci.edu/ml/datasets/Molecular+Biology+(Splice-junction+Gene+Sequences)}{DNA}                 }
\item{\href{https://archive.ics.uci.edu/ml/datasets/glass+identification}{Glass Identification}                 }
\item{\href{http://archive.ics.uci.edu/ml/datasets/madelon}{Mandelon} \cite{madelon_data}                       }
\item{\href{https://archive.ics.uci.edu/ml/datasets/soybean+(small)}{Soybean} \cite{soybean_data}                }
\item{\href{https://archive.ics.uci.edu/ml/datasets/Yeast}{Yeast} \cite{yeast_data}                             }
\end{itemize}
\end{multicols}

Our TreeGrad is based on a two stage process. First, constructing a tree where the decision boundaries are oblique boundaries. Next, sparsifying the neural network with axis-parallel boundaries and fine tuning the decision tree. We consider the usage of $\ell_0$ regularizer combined with $\ell_1$ regularizer in a manner described by \cite{louizos2017learning}. We found sparsifying neural networks preemptively using the $\ell_0$ regularizer enabled minimal loss in performance after neural networks were compressed to produce \textit{axis-parallel} splits. 

All trees were grown with the same hyperparameters with maximum number of leaves being set to $32$. The base architecture chosen for TreeGrad algorithm was determined by LightGBM, where the results shown in tables \ref{tbl-featimp1} and \ref{tbl-dt} are when all weights are re-initialise to random values. For our boosted trees, we use an ensemble of 100 trees for all datasets and boosted tree implementations.

\begin{table}
  \caption{Kendall's Tau of TreeGrad and LGM and GBT Feature Importance (Split Metric). Larger values mean `more similar'.}
  \label{tbl-featimp1}
  \centering
\begin{tabular}{|l|rr|rr|}
\hline
         & \multicolumn{2}{c|}{Decision Tree}                 & \multicolumn{2}{c|}{Boosted Trees (100 Trees)}         \\
         & \multicolumn{1}{r}{LGM} & \multicolumn{1}{r|}{GBT} & \multicolumn{1}{r}{LGM} & \multicolumn{1}{r|}{GBT} \\ \hline
    adult     & -0.10   & -0.33 & 0.48 & 0.47 \\
    covtype   & 0.22    & 0.33 & 0.45 & 0.44 \\
    dna       & 0.12   & 0.13 & -0.06 & 0.28 \\
    glass     & 0.34    & 0.07 & 0.17 & 0.11 \\
    mandelon  & 0.54    & 0.59 & 0.05 & 0.07 \\
    soybean   & 0.08    & 0.21 & 0.05 & 0.05 \\
    yeast     & 0.47  & -0.28 & 0.47 & 0.6 \\   \hline
\end{tabular}
\end{table}

When comparing the diversity in features which are selected, we notice that greedy methods, being LGM and GBT generally select similar kinds of features. However, in some instances, our non-greedy approach selectes different kinds of features which can be observed in the \textbf{adult} data set when considering a single decision tree as shown in Table \ref{tbl-featimp1}. When we consider the boosting approaches, as there are more trees which are build, and hence more split candidates, the level of agreement changes in distribution. What is interesting is the three datasets with the lowest amount of agreement (DNA, mandelon, soybean) with our TreeGrad approach actually performs the best against the test dataset as shown in Table \ref{tbl-dt} which suggests that our TreeGrad algorithm has managed to find some relationships which greedy approaches may have missed. 

\begin{table}
  \caption{Accuracy Performance of TreeGrad against LGM and GBT (Test Dataset). Larger values means better performance.}
  \label{tbl-dt}
  \centering
\begin{tabular}{|l|lrr|lrr|}
\hline
                     & \multicolumn{3}{l|}{Decision Tree}                            & \multicolumn{3}{l|}{Boosted Trees (100 Trees)}                            \\
Dataset              & TreeGrad & \multicolumn{1}{c}{LGM} & \multicolumn{1}{c|}{GBT} & TreeGrad & \multicolumn{1}{c}{LGM} & \multicolumn{1}{c|}{GBT} \\ \hline
    adult     &   \textbf{0.797} &    0.765 &    0.759    &     0.860  &  0.873 &   \textbf{0.874}  \\       
    covtype   &   0.644 &    \textbf{0.731} &    0.703    &     0.832 & \textbf{0.835} & 0.826 \\            
    dna       &   0.850 & 0.541 & \textbf{0.891}          &     \textbf{0.950} & 0.949 & 0.946 \\            
    glass     &   \textbf{0.688} &   0.422 &  0.594       &     0.766 & \textbf{0.813} & 0.719 \\            
    mandelon  &   \textbf{0.789} &   0.752 &   0.766      &     \textbf{0.882} & 0.881 & 0.866 \\            
    soybean   &   0.662 &    0.583 &    \textbf{0.892}    &     \textbf{0.936} & \textbf{0.936} & 0.917 \\   
    yeast     &   \textbf{0.553} &  0.364 &  0.517        &     \textbf{0.591} & 0.573 & 0.542 \\            
\hline
Number of wins       & \textbf{4} & 1 & 2 & \textbf{4} & 3 & 1 \\
Mean Reciprocal Rank & \textbf{0.762} & 0.452 & 0.619 & \textbf{0.762} & 0.714 & 0.429     \\ 
\hline
\end{tabular}
\end{table}

We also compare the comparison of all three approaches against the test datasets as shown in Table \ref{tbl-dt}. Overall the best performing model is our approach (TreeGrad). TreeGrad's performance is a demonstration that non-greedy approaches can fall back to representations similar to greedy approaches which is competitive with existing greedy algorithms, or discover representations which can not be recovered by greedy approaches. We observe that for boosting approaches LGM has very similar performance compared with TreeGrad, whereas GBT performance begins to fall off compared with its strong performance when considering the single decision tree.




\section{Conclusion}

We have demonstrated approaches to unify boosted tree models and neural networks, allowing tree models to be transferred to neural network structures. We have provided an approach to rebuild trees in a non-greedy manner and decision splits in the scenario were weights are reset and provided learning bounds for this approach. This approach is demonstrated to be competitive and even an improvement over current greedy tree ensemble algorithms like LightGBM, and empirically better than popular frameworks like Scikit-learn. 

There are several avenues for future work. We want to extend this approach to allow decision trees to induced in an online manner; we want to extend the layers within a decision tree to include the ability to use conventional CNN to create new architectures over richer sources of data which decision trees are not usually trained against.

\bibliographystyle{splncs04}
\bibliography{main}

\end{document}